\documentclass[conference]{IEEEtran}
\IEEEoverridecommandlockouts
\usepackage{cite}
\usepackage{graphicx}
\usepackage{amsmath}
\usepackage{amssymb}
\usepackage{booktabs}
\usepackage{subcaption}

\usepackage{tabularx}
\usepackage{adjustbox}
	
\usepackage{color, colortbl}
\usepackage[normalem]{ulem}

\usepackage[natural]{xcolor}
\usepackage[markup=underlined]{changes}
\usepackage{todonotes}
\usepackage{mathtools}
\usepackage{url}
\usepackage{algorithm}
\usepackage{algorithmicx}
\usepackage[noend]{algpseudocode}
\usepackage{gensymb}
\usepackage{siunitx} 
\usepackage{interval}
\usepackage{tikz}
\usepackage[pagebackref,breaklinks,colorlinks]{hyperref}

\newcommand\copyrighttext{%
  \footnotesize \textcopyright 2023 IEEE. Personal use of this material is permitted.
  Permission from IEEE must be obtained for all other uses, in any current or future
  media, including reprinting/republishing this material for advertising or promotional
  purposes, creating new collective works, for resale or redistribution to servers or
  lists, or reuse of any copyrighted component of this work in other works.}

\newcommand\copyrightnotice{%
\begin{tikzpicture}[remember picture,overlay]
\node[anchor=south,yshift=10pt] at (current page.south) {\fbox{\parbox{\dimexpr\textwidth-\fboxsep-\fboxrule\relax}{\copyrighttext}}};
\end{tikzpicture}%
}

\title{Sensor Equivariance by LiDAR Projection Images}

\author{Hannes Reichert, Manuel Hetzel, Steven Schreck, Konrad Doll, and Bernhard Sick
	\thanks{H. Reichert, M. Hetzel, S. Schreck, and K. Doll are with the Faculty of Engineering,
		University of Applied Sciences Aschaffenburg, Aschaffenburg, Germany
		{\tt\footnotesize hannes.reichert@th-ab.de,
			manuel.hetzel@th-ab.de, steven.schreck@th-ab.de konrad.doll@th-ab.de}}
	\thanks{B. Sick is with the Intelligent Embedded Systems Lab, University of Kassel,
		Kassel, Germany
		{\tt\footnotesize bsick@uni-kassel.de}}
}

\begin{document}

\maketitle
\copyrightnotice

\begin{abstract}
In this work, we propose an extension of conventional image data by an additional channel in which the associated projection properties are encoded. This addresses the issue of sensor-dependent object representation in projection-based sensors, such as LiDAR, which can lead to distorted physical and geometric properties due to variations in sensor resolution and field of view. To that end, we propose an architecture for processing this data in an instance segmentation framework. We focus specifically on LiDAR as a key sensor modality for machine vision tasks and highly automated driving (HAD). Through an experimental setup in a controlled synthetic environment, we identify a bias on sensor resolution and field of view and demonstrate that our proposed method can reduce said bias for the task of LiDAR instance segmentation. Furthermore, we define our method such that it can be applied to other projection-based sensors, such as cameras. To promote transparency, we make our code and dataset publicly available. This method shows the potential to improve performance and robustness in various machine vision tasks that utilize projection-based sensors.

\begin{figure}[!h] 
    \centering
    \includegraphics[width=\columnwidth]{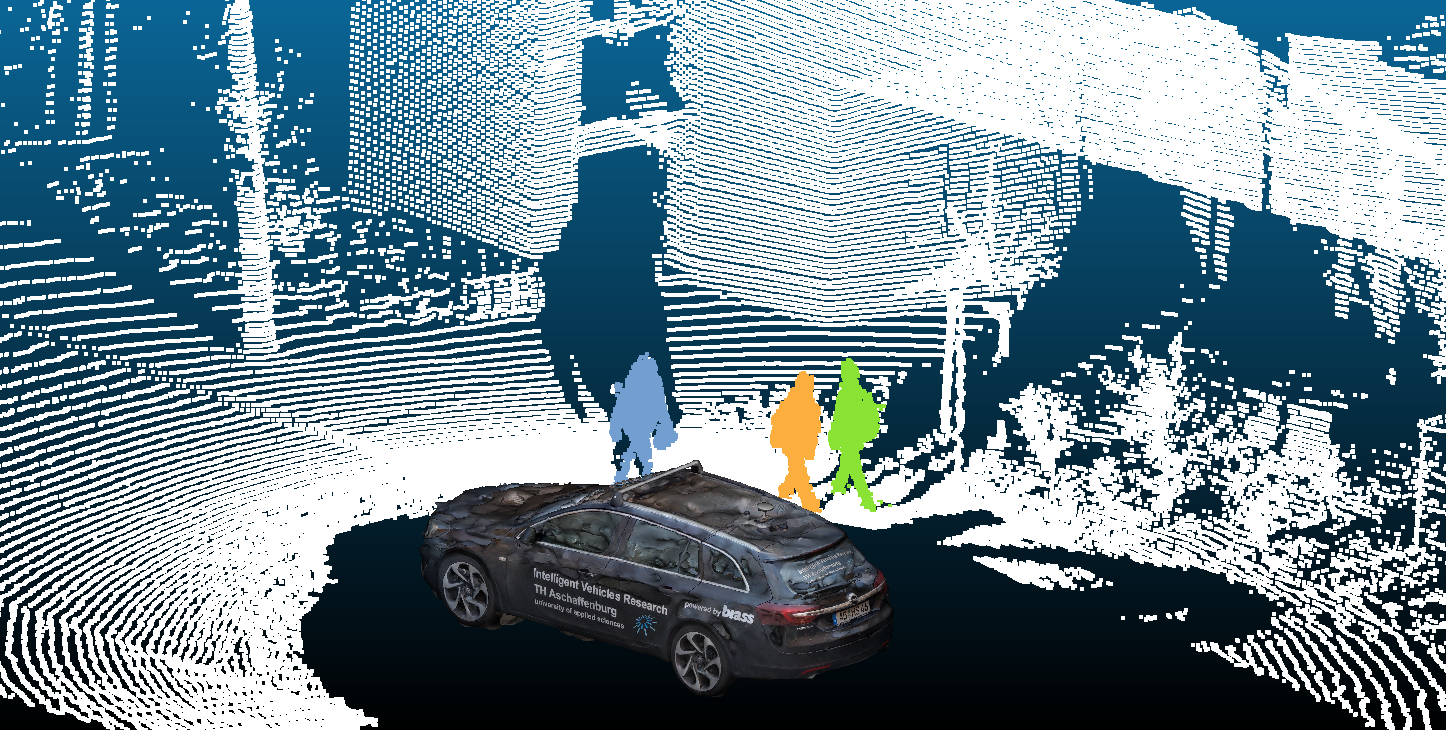}
    \caption{\textit{\textbf{Visual Abstract}}: Instance segmentation of a LiDAR scan.}
    \label{fig:visual_abstract}
\end{figure}

\end{abstract}
\section{\large Introduction}
\label{sec_introduction}

This work aims to address the problem of generalizing a machine learning model, specifically instance segmentation, trained on data from a single sensor or a collection of sensors, to new sensors with different characteristics, such as field of view and resolution. The motivation behind this research is the fact that a wide variety of sensors with different properties are integrated into many products, and the functionalities and capabilities of these products heavily depend on the sensors used. This means that new sensors will be continuously introduced as they become available.

One important application for this research is in the field of autonomous vehicles, specifically in creating a scene understanding from sensor data using segmentation, which assigns a class and instance label to each data point, such as a 3D point of a LiDAR scan or a pixel of a camera image as shown in \autoref{fig:visual_abstract}. Autonomous driving perception modules usually consist of data-driven models based on sensor data, but these models may be biased toward the sensor used for data acquisition, which can seriously impair the transferability of the perception models to new sensor setups.

For LiDAR sensors, in particular, numerous manufacturers have emerged in recent years, adding new technologies and sensors to the market. Various previous approaches have projected a LiDAR 3D point cloud onto a 2D spherical range image using efficient 2D convolutional operations and architectures for image segmentation. We propose using deflection metric to solve the problem of transferring machine learning model from a single sensor to new sensors.

\subsection{Problem Statement and Intuition} \label{sec:problem}
\begin{figure}
     \centering
     \begin{minipage}{0.49\columnwidth}
        \includegraphics[width=\linewidth, height=4.5cm]{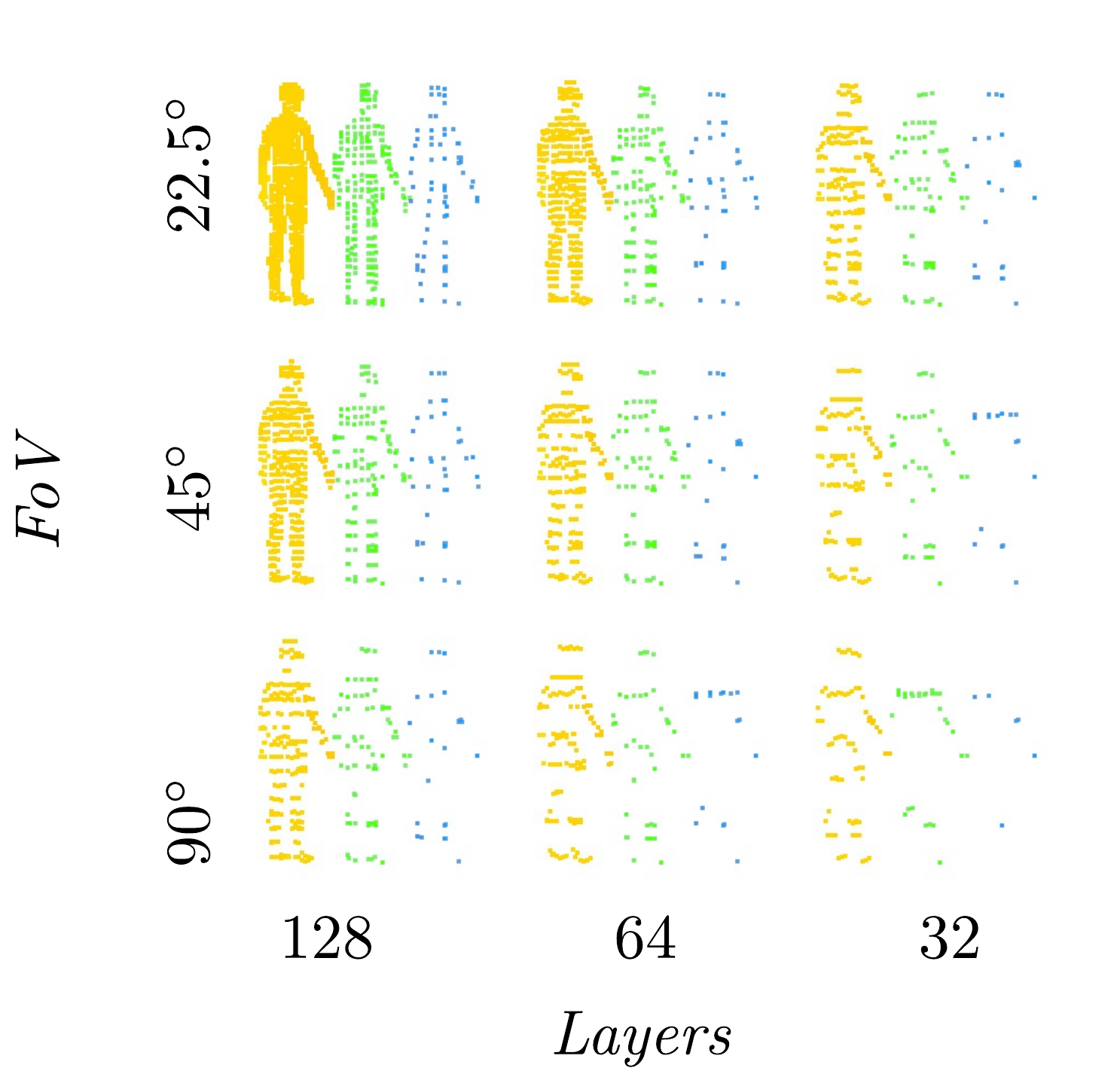}
        \subcaption{LiDAR Scan.}
    \end{minipage}
    \hfill
    \begin{minipage}{0.49\columnwidth}
        \includegraphics[width=\linewidth, height=4.5cm]{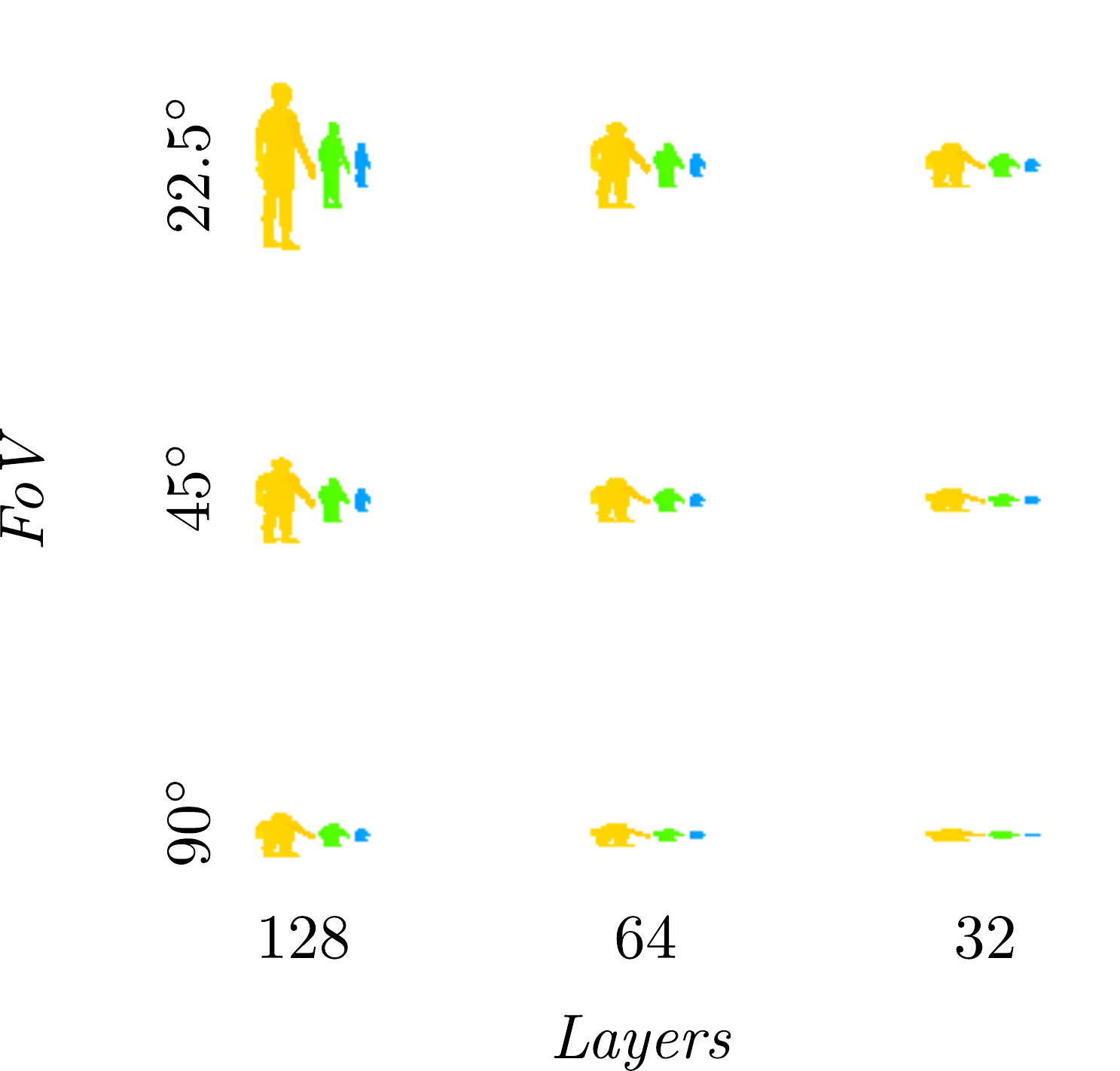}
        \subcaption{Spherical Projection.}
    \end{minipage}
    \caption{Simulated LiDAR scans of a pedestrian at different fields of view and vertical resolutions at the distances 10 m, 20 m, and 40 m, in yellow, green, and blue, respectively (left). Spherical projections of those scans (right).}
    \label{fig:problem}
\end{figure}
LiDAR point clouds are unstructured and sparse. As the distance between the sensor and the object increases, the number of measurement points on the object decreases, making it difficult for machine learning models to accurately detect and classify the object. Mainly because it's hard to find connected components in an unstructured point cloud at different levels of sparsity. The level of sparsity is further compounded by the sensor's Field of View (FoV), vertical resolution, and azimuth resolution. In \autoref{fig:problem} this is shown for a person distanced 10 m, 20 m, and 40 m from sensors with differences in resolution and field of view. 

With spherical projections of LiDAR scans are considered. Instead of becoming sparse, the size of the object changes in pixel space \autoref{fig:problem}. Thus, relaxing the issue of dealing with lack of structure and sparsity to the issue of dealing with different scales in pixel space. 

\subsection{Related Work} \label{sec_sota}

For image data, there is some preliminary work, addressing the problems stated in \autoref{sec:problem}:
Liu et al. introduce CoordConvs \cite{CoordConvs}, as a solution for the coordinate transform problem.
The convolution itself is spatial invariant. However, for some tasks, spatial variance might be needed. In object detection, for example, the coordinate transform problem arises from processing features in pixel space and output bounding boxes in cartesian space. 
A CoordConv layer is a simple extension of the standard convolutional layer. It has the same functional signature as a convolutional layer, but utilizes extra channels for the incoming representation. These channels contain hard-coded coordinates, the most basic version of which is one channel for the $u$ coordinate and one for the $v$ coordinate. The authors claim that the CoordConv layer keeps the properties of few parameters and efficient computation from convolutions, but allows the network to learn if spatial variance or invariance is needed for learning the task. This is useful for coordinating transform-based tasks where regular convolutions can fail. However, the authors proved their concept on simplified tasks.

Wang et al. \cite{wang2021solo}\cite{wang2020solov2} utilize the CoordConv concept as a component in an instance segmentation framework. The authors argue that spatially variant convolutions are necessary for instance segmentation, which is related to semantic segmentation. Furthermore, they concluded that few CoordConv layers within the backbone are enough to achieve this.

Facil et al. teach camera-aware multi-scale convolutions for depth estimation from image data \cite{CamConv} supplied to a neural network. The method comprises pre-computing pixel-wise coordinates and horizontal and vertical field-of-view maps, fed with input features to a convolution operation. These maps are supplied to the neural network with different resolutions and on different layers to allow the network to learn and predict depth patterns that depend on the camera calibration. The authors conclude that the neural network supplied with the respective maps can generalize over camera intrinsics and allow depth prediction networks to be camera-independent.
This work can be seen as the work being most closely related to our approach. Even if this method can be adapted for image data in general, it utilizes four additional channels exclusively designed for camera sensors.

In order to process LiDAR data with such a method, the data must first be converted into an image representation.
Many works exist that convert LiDAR point clouds into images through a spherical projection. The SemanticKITTI benchmark \cite{behley2019iccv} and RangeNet \cite{8967762} have contributed significantly to this development. The elegance of spherical projection is that LiDAR data can be processed similarly to camera data with a Convolutional Neural Network (CNN). This also includes a significant improvement in runtime when processing LiDAR data. Methods such as Lite-HDSeg \cite{litehdseg}, SalsaNext \cite{SalsaNet}, or SqueezeSeg \cite{SqueezeSegV2} all consider spherical projections and propose different architectures for processing.  

In \cite{LiDARNet} the authors present a boundary-aware domain adaptation
model for LiDAR scan full-scene semantic segmentation. Their method considers boundary information while learning to predict full-scene semantic segmentation labels. They also use spherical projection of LiDAR data and demonstrate the adaption of their model to different sensors. They mainly use sensors with similar resolutions. In this respect, they do not address the problems we state in \autoref{sec:problem}. 

\subsection{Main Contributions} \label{sec_contrib}
This paper presents a novel approach for encoding the projection properties of a sensor in an image representation called the deflection metric. The deflection metric is a one-channel image that can resolve ambiguities in the projection and homogenize data from various sensors.

To demonstrate the effectiveness of the deflection metric, we propose a backbone architecture based on Faster-RCNN and a spherical projection model for spinning LiDAR sensors, adapted from the pinhole camera model. Combined with range information the deflection metric can define a 3D coordinate system suitable for processing with CNNs.

We conducted experiments using a high-resolution LiDAR dataset generated with the CARLA \cite{Dosovitskiy17} simulator. To evaluate the transfer ability of our method and conduct an ablation study to confirm its usability on sensors with different fields of view.

For transparency and reproducibility, we also included the source code for the data generation process, training and evaluation of our models, and a collection of pre-trained models.

\section{\large Method}
\label{sec_method}
Our method aims to address the issue of ambiguity and distortion in images by providing additional knowledge about the sensor used to capture the image. This additional information is encoded in a deflection metric, which can be interpreted or processed by a CNN.

First, we define the deflection metric, which encodes the projection properties of a sensor, such as the sensor's field of view and distortion. This is done by creating a spherical projection of the sensor data, and transforming it into an image representation (see sections \autoref{sec:spherical projection} and \ref{sec:deflection}).

Second, we show how the deflection metric is provided as input to a state-of-the-art object detection and instance segmentation architecture (see \autoref{sec:backbone}), which enables the CNN to better understand the image, reducing the ambiguities in scale and distortions.

\subsection{Spherical Projection}
\label{sec:spherical projection}
There has been a growing interest in using spherical projections for LiDAR instance segmentation in recent years. Spherical projections offer several advantages, such as being able to capture the full 360-degree field of view and being able to maintain consistent point density across the whole projection.

\begin{figure}[!h] 
    \centering
    \includegraphics[width=\columnwidth]{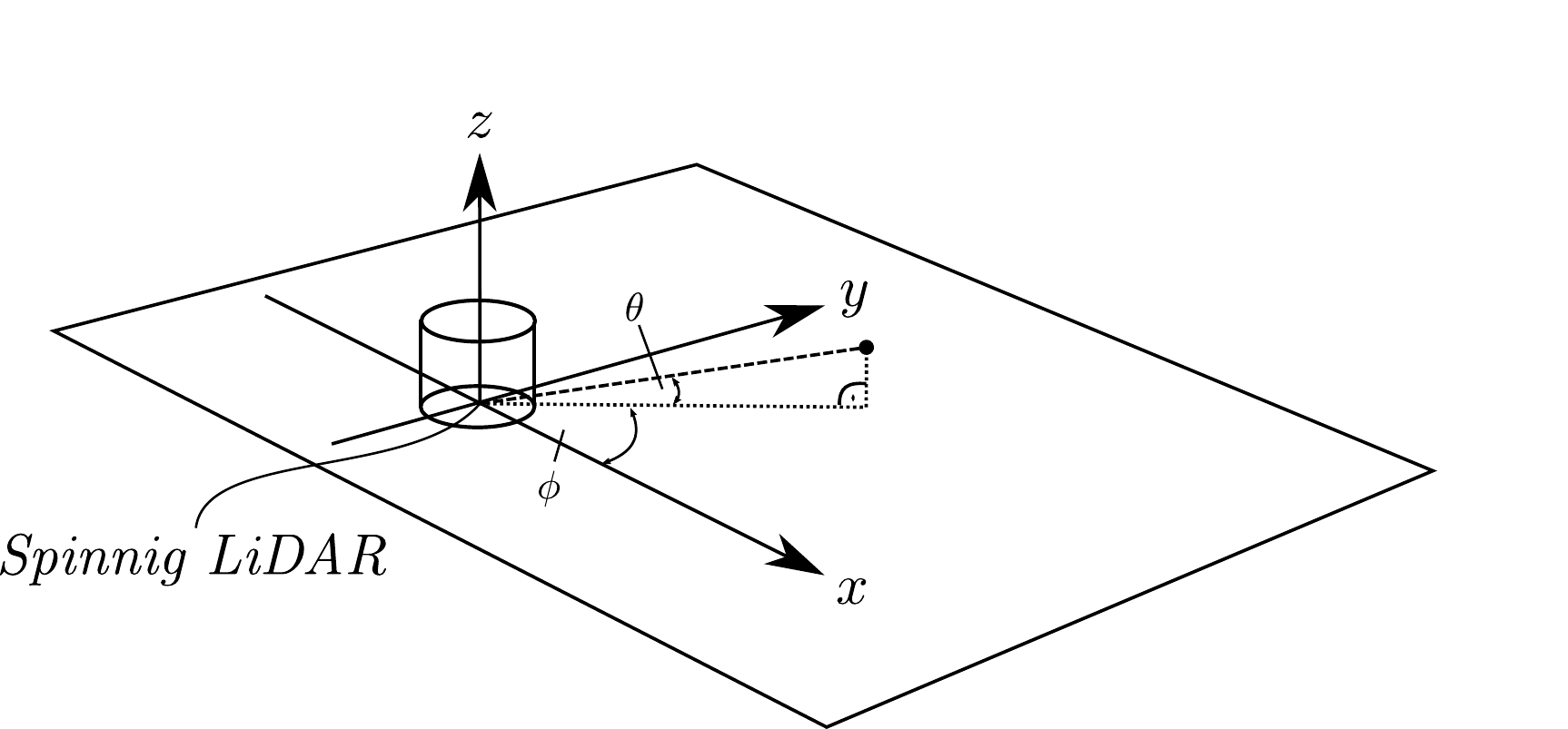}
    \caption{Coordinate system of a spinning LiDAR. $\phi$ as the azimuth angle and $\theta$ as an inclination angle from the $xy$-plane.}
    \label{fig:spinning_lidar}
\end{figure}

\begin{figure}[!h] 
    \centering
    \includegraphics[width=\columnwidth]{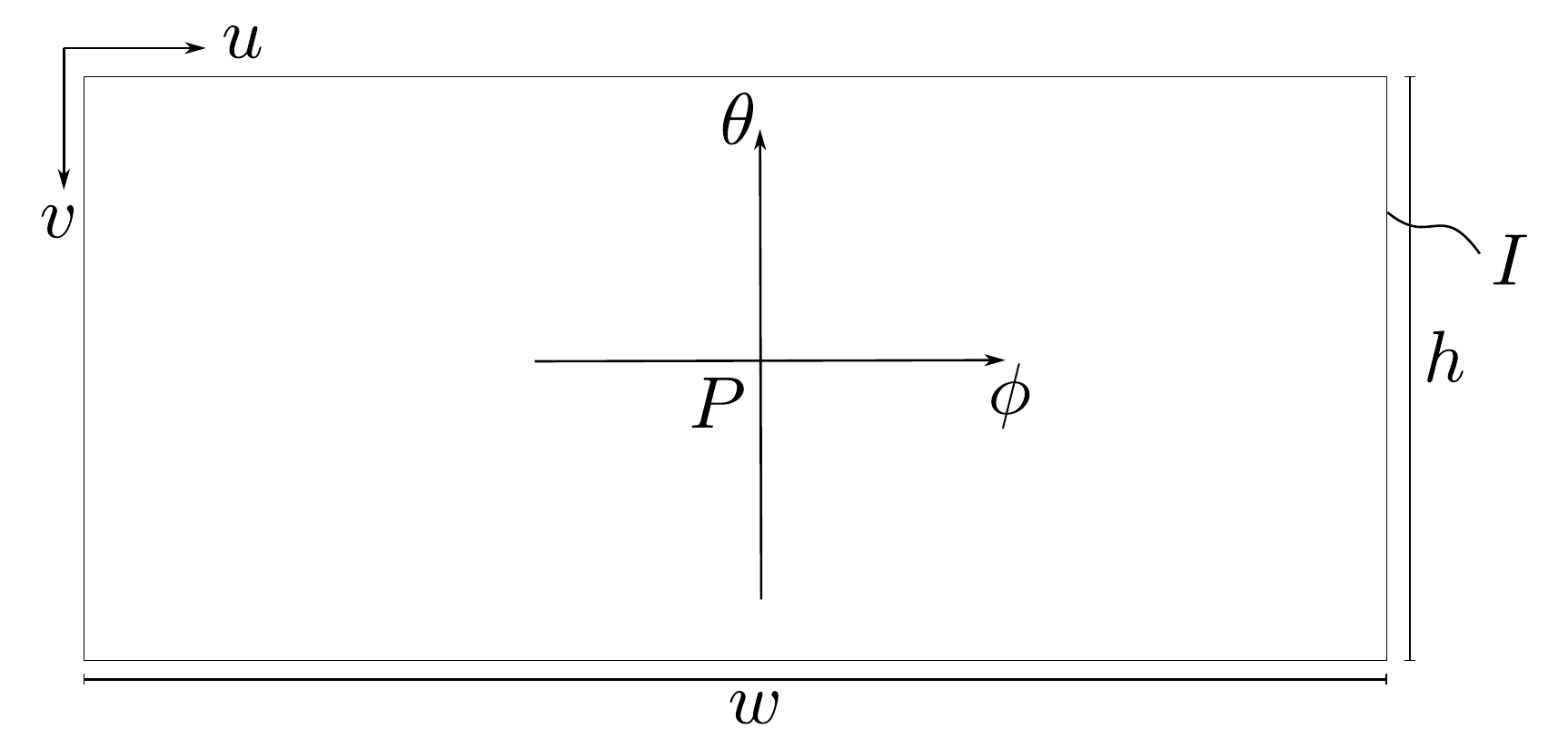}
    \caption{Spherical image $I$ with principal point $P$, height $h$, width $w$.}
    \label{fig:spherical}
\end{figure}

The process of projecting point clouds from a spinning LiDAR sensor into a spherical image involves converting the Cartesian coordinates of the measurement points into spherical coordinates. Specifically, the process converts each point in the point cloud represented by its Cartesian coordinates $[x,y,z]^T$, into spherical coordinates represented by $[phi, theta, r]^T$. This conversion is illustrated in Fig. \ref{fig:spinning_lidar}. The phi coordinate corresponds to the angle of the point in the XY plane, theta is the angle from the positive Z-axis, and r is the distance from the origin.
This spherical projection is a way to capture the geometry of the sensor in a single image. Subsequently, we use the following projection model to generate a spherical range image:
\begin{equation}
\underbrace{\begin{bmatrix}
 u\\
 v\\
 1\\
\end{bmatrix} }_{\vec{u}}
=
\underbrace{\begin{bmatrix}
\frac{1}{\triangle \phi} &  0 &  c_{\phi}\\
 0 &  \frac{1}{\triangle \theta} &  c_{\theta}\\
 0 & 0 & 1\\
\end{bmatrix}}_{\mathbb{K}} \cdot \underbrace{\begin{bmatrix}
 \phi\\
 \theta\\
 1\\
\end{bmatrix}}_{\vec{x}}  
\end{equation}
Analogous to the projection model of pinhole cameras, the projection matrix $\mathbb{K}$ describes a discretization $\triangle \phi, \triangle \theta$ along the angles $\phi, \theta$ and a shift of the center coordinates $c_{\phi}, c_{\theta}$ defined by the height and width of the resulting image. Since the discretization can cause several points to be projected onto one pixel, we only use the points with the smallest Euclidean distance $r$ to the sensor. For a conventional spinning LiDAR sensor, the image height $h$ and width $w$ will be equivalent to the number of layers and azimuth increments, respectively, as displayed in Fig.~\ref{fig:spherical}.

With the spherical projection, an image representation $I$ can be constructed. Points from a 3D point cloud and auxiliary data can be projected to this ordered image representation. They result in several images for a LiDAR scan.  E.g. $I_{r}$ for the Euclidean distance.

\subsection{Deflection Metric}
\label{sec:deflection}
\begin{figure}[!h] 
    \centering
    \includegraphics[width=\columnwidth]{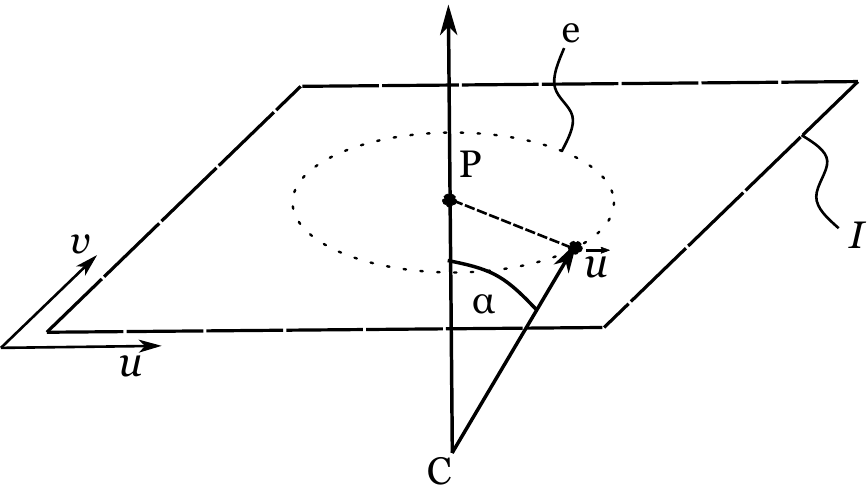}
    \caption{Deflection metric $\alpha$: is calculated as an angle between the optical axis, defined by a sensor's origin $C$ and a principal point $P$, and a pixel $\vec{u}=[u,v]^T$.}
    \label{fig:sens_dat_abs}
\end{figure}
With the deflection metric, as shown in \autoref{fig:sens_dat_abs}, we propose a method to encode the geometric characteristics of a sensor alongside its data. We do this using a deflection image $I_{\alpha}$, which encodes an inclination angle $\alpha$ in every pixel as a one channel image. $\alpha$ provides a consistent relation between a projected 3D point and the position of this 3D point in relation to the sensor and is, therefore, compatible among sensors with different characteristics. The deflection metric $\alpha$ is defined as an angle between the optical axis, defined by a sensor's origin $C$ and a principal point $P$, and a pixel $\vec{u}=[u,v]^T$. $\alpha$ essentially describes half the field of view at a pixel position $\vec{u}$. 

The projection can be modeled by a sensor model and parameterized by linear or nonlinear sensor intrinsics. This intrinsics can be calculated based on the constituents of the projection system, e.g., the position and properties of lenses and apertures, measured in an optical setup or estimated by calibration. 

Based on a parameterized sensor model, the deflection metric  $\alpha$ for a pixel position $\vec{u}$ can be determined by the sensor's intrinsic matrix $\mathbb{K}$. Based on the image coordinates, $\alpha$ of a pixel $\vec{u}$ with respect to the projection center $P$ can be determined:
\begin{equation}
\alpha(\vec{u}) = \sqrt{\left[\mathbb{K}^{-1}\vec{u}\right]^T\left[\mathbb{K}^{-1}\vec{u}\right] - 1}
\label{equ:deflection}
\end{equation}
All the same values of the deflection metric $\alpha$ are arranged on a circle or ellipse $e$ (see dashed circle in \autoref{fig:sens_dat_abs}) in image data. 
From $\alpha(\vec{u})$ a normed distance $d(\vec{u})$, such that the distance between sensor center $C$ and principal point $P$ is one, can be calculated by $d(\vec{u})= tan(\alpha(\vec{u}))$. We would like to emphasize that due to the change of coordinates to spherical coordinates, \autoref{equ:deflection} does not calculate the $\alpha(\vec{u})$ when using a camera intrinsic matrix $\mathbb{K}$, but the normalized distance $d(\vec{u})$. To obtain $\alpha(\vec{u})$ for a camera sensor, the inverse trigonometric function $\alpha(\vec{u})= arctan(d(\vec{u}))$ has to be used.

The deflection image $I_{\alpha}$ is a one-channel image, in which every pixel $\vec{u}$ is aligned with the data image $I$. This increases the information content of each pixel by the geometric sensor properties. The deflection image $I_{\alpha}$ can be processed together with the sensor data by convolutional layers of a CNN. Unlike image data, the deflection image is not invariant to translation, rotation, and scale. However, since the convolutional layers of CNNs are learned, a CNN can decide whether to use this additional information in the learning process. 
The range image $I_{r}$ and the deflection image $I_{\alpha}$ are components of a 3D coordinate system and designed to be processable with convolutional filters. 

\subsection{Top-Down Injection}
\label{sec:backbone}

\begin{figure}[!h] 
    \centering
    \includegraphics[width=\columnwidth]{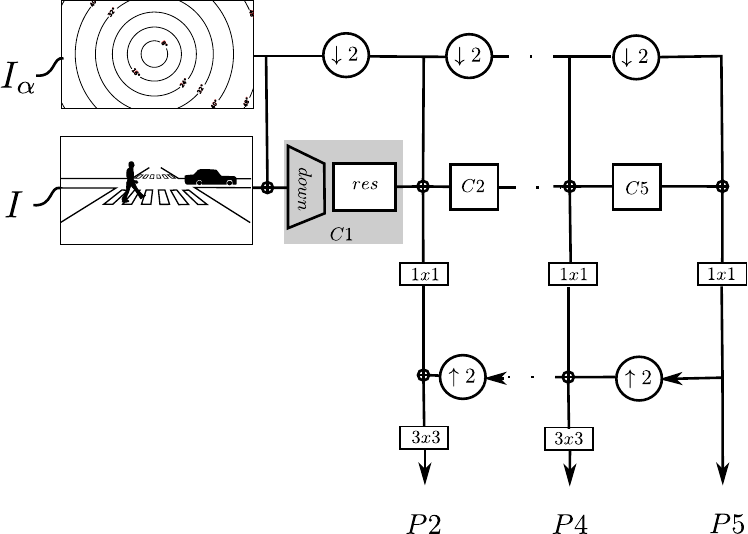}
    \caption{Backbone Architecture. The deflection metric is injected before every pyramid stage. $\downarrow2$ denotes a down-sample operation, $\uparrow2$ denotes an up-sample operation, $\bigoplus$ denotes a channel wise concatenation.}
    \label{fig:architecture}
\end{figure}
An existing backbone meta-architecture is modified to inject the deflection metric into the model at the input and selected locations. Based on the findings in \cite{wang2021solo}\cite{wang2020solov2} we decided to use a ResNet50-FPN. The modification to the ResNet-FPN is shown in \autoref{fig:architecture}. At the input stage, the deflection image $I_{\alpha}$ is concatenated with a three-channel image $I$, resulting in an input shape of $h\times w \times 4$. The image input is processed top-down in five down-sampling stages. Each stage halves the height $h$ and width $w$. This is done for the image data using strided convolutions, followed by a residual block ($C1$ to $C5$). The deflection image $I_{\alpha}$ is down-sampled in parallel and concatenated to the features of the stages $C1$ to $C5$. This ensures that the feature map can be used in every stage. After the injection, a $1\times1$-convolution is used to fuse the features of the stage with the deflection metric. This allows the network to keep or discard the deflection metric for a particular stage. The feature maps are up-sampled from the bottom up prior to a fusion with the pristine feature map from the respective stages. The fusion is performed by a channel-wise concatenation of the feature maps and a subsequent $3\times3$-convolution for anti-aliasing, as with common FPN architectures. This results in the pyramid stages $P$ with the respective shapes $(h/2^{i})\times( w/2^{i})\times256$ ($i$ denotes the stage index). The pyramid stages are then fed into a semantic segmentation head, as described in \cite{LinFPN}.

\subsection{Data Augmentation}
This section discusses the use of image augmentation in the training process of the proposed method. Image augmentation is a technique of altering the existing data to create more data for the model, which is especially important when the training data comes from a single sensor source, but the model should work sensor equivariant.

The proposed method uses geometric image-based operations to augment both the range image and the deflection image. The operations used are resize and center-crop. The resize operation changes the resolution of the sensor, and the center-crop operation changes the field of view. The combination of both allows the simulation of various sensors during training. It's worth noting that with this kind of augmentation, only sensors with an equal or smaller field of view and resolution can be simulated.

In the context of LiDAR semantic segmentation, the augmentation process modifies the vertical and azimuthal resolution of the sensor, as well as the vertical field of view. This helps the model learn to generalize to new sensor configurations.


\section{\large Experiment}
\label{sec:expirement}

\subsection{Dataset}

\textbf{Data Generation}
We used the CARLA simulator for data generation. In CARLA we simulated an ultra-high resolution LiDAR with full dome coverage, mounted to a carrier vehicle, resulting in dense point clouds with semantic segmentations, instance segmentations, cuboids, and meta information. 
For the instance segmentation, we consider three classes of road user $C \in \{\textit{pedestrian},\textit{vehicle},\textit{two-wheeler}\}$. Our simulated high-resolution LiDAR is beyond current commercially available sensors in terms of resolution and FoV. However, we can generate LiDAR sensors with more reasonable parameters regarding resolution and field of view from the high-resolution LiDAR, while maintaining initial conditions like semantic context, giving us a controlled environment for our experiment. 
We generate our simulated LiDAR frames at a frame rate of 1 Hz so that two consecutive frames do not look too similar. We also chunked our dataset into scenes of 100 frames. For each scene, we altered the simulation location (i. e. map), the initial parameters for traffic simulation (e.g. amount, type, and behavior of other road users), the carrier vehicle, and its trajectory. This totals 24 sequences, from which we use 15 as training data and 9 for testing. In addition, we ensured that the carrier vehicle and the simulation location differed in the test and train sequences.
The high-resolution LiDAR scans consist of a vertical FoV of $f_v = 180^\circ$, a horizontal FoV of $f_h = 360^\circ$, the number of azimuth increments $w=2048$, and the number of horizontal layers $h=1024$. This results in a spherical projected range image $I_r$ (as described in \autoref{sec:spherical projection}) with a resolution $1024\times2048$.

\begin{figure}[h]
     \centering

    \begin{minipage}{\columnwidth}
        \centering
        \includegraphics[width=0.8\linewidth]{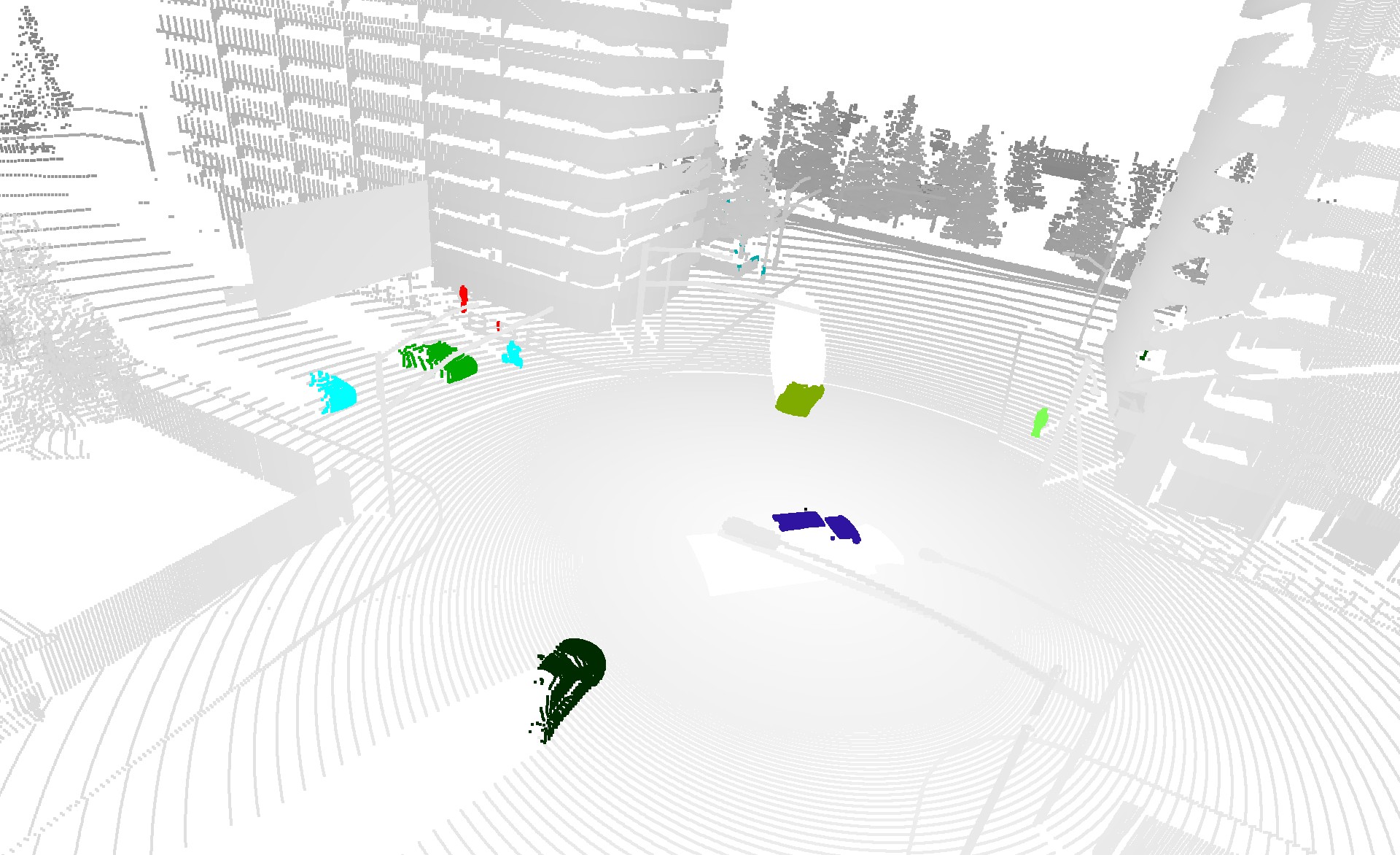}
        \subcaption{Point Cloud.}
    \end{minipage}
    \hfill
    \begin{minipage}{\columnwidth}
        \centering
        \includegraphics[width=0.8\linewidth]{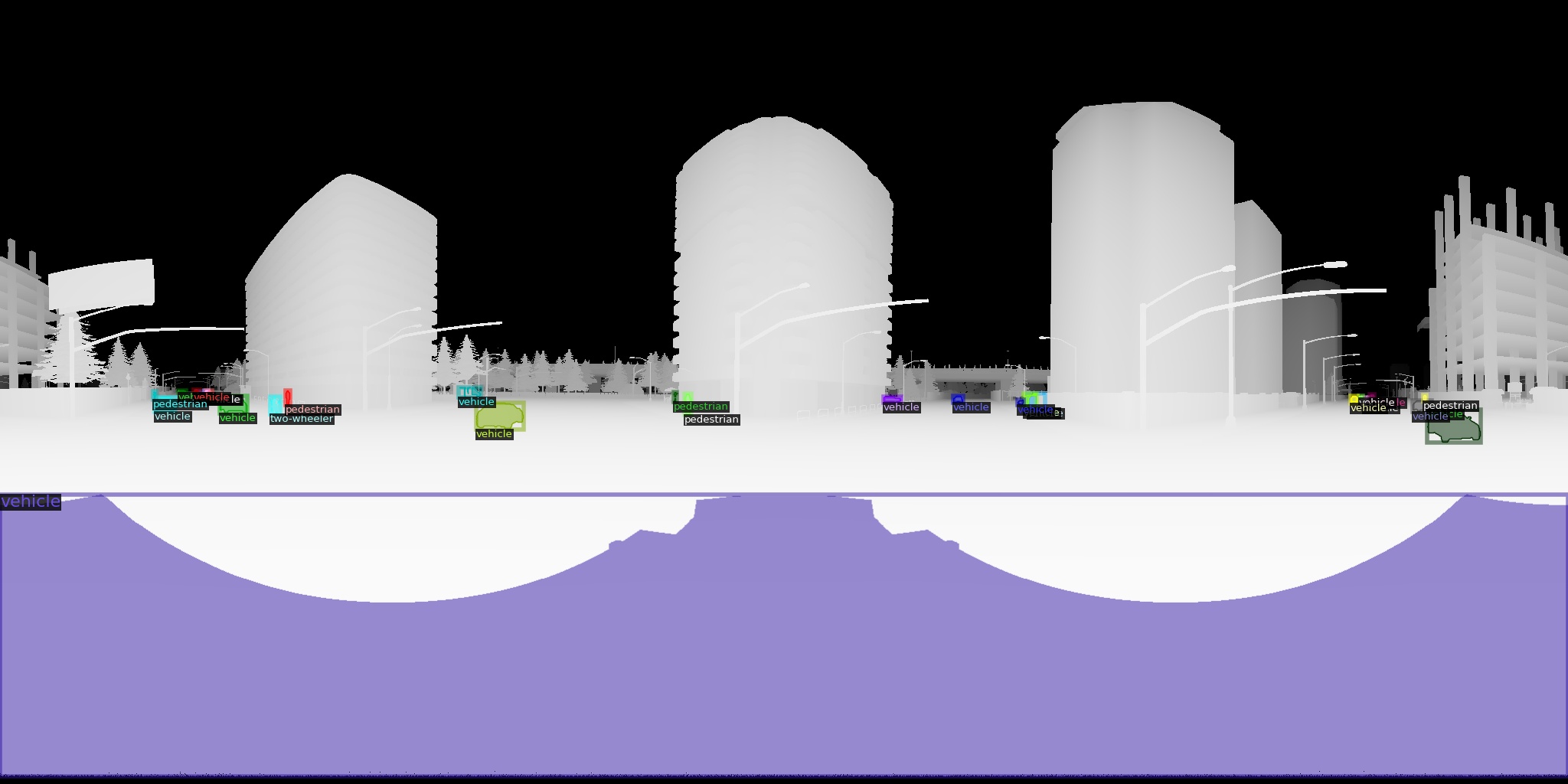}
        \subcaption{Spherical Projection.}
    \end{minipage}
    \caption{Example frame with instance labels of the simulated high-resolution LiDAR.}
    \label{fig:dataset}
\end{figure}

\subsection{Setup}
\textbf{Implementation details}: We implemented our method in Detectron2 \cite{wu2019detectron2} and used their reference implementations. For training, we used a single NVIDIA 3090 TI GPU. For the instance segmentation, we use  MaskRCNN \cite{maskRCNN} and PointRend \cite{kirillov2019pointrend} as application head. For the backbone, we modified a ResNet-50 FPN as described in \autoref{sec:backbone}. All models were trained until convergence. 
\\

\textbf{Evaluation and Metrics}:
We use mean average recall (mAR) as defined in COCO. We prefer mAR over mean average precision (mAP) since it indicates how many objects are correctly detected. A higher recall is preferable for downstream tasks like object tracking since removing false positive detection over time is easier than adding missed detections. 
\\

\textbf{Experimental Setup}: \label{sec:experiment_B}
In the following, we describe our experimental design to test our method's generalizability to sensors with other characteristics (e.g. FoV and resolution). We projected the high resolution LiDAR scans and instance labels to sensors with different characteristics. We mainly changed the number of azimuth increments, the number of horizontal layers, and the vertical FoV to $w \in \{512,1024,2048\}$, $h \in \{32,64,128\}$, and  $f_v \in \{22.5^\circ, 45^\circ, 90^\circ\}$ respectively. We use the spherical projection described in \autoref{sec:spherical projection}  to project the point clouds to the image representation $I$. We then resize the projected image $I$ accordingly to $h$ and $w$. 
In this way, we reproduce sensors that are currently on the market. Currently, spinning LiDAR with a vertical resolution of 128 and a horizontal resolution of 2048 from manufacturers such as Ouster or Velodyne represent the commercial state of the art. However, these are currently still expensive, and if this aspect and also the required runtime are taken into account, sensors with lower resolution are also relevant for research and industrial applications. However, sensors with a lower resolution can be augmented from these high-resolution sensors at any time. This is not possible with FoV. For real world data acquisition, we recommend to use high resolution sensors if possible.

\subsection{Ablation Study}

\begin{figure*}[h]
    \centering
    \includegraphics[clip,width=0.8\textwidth]{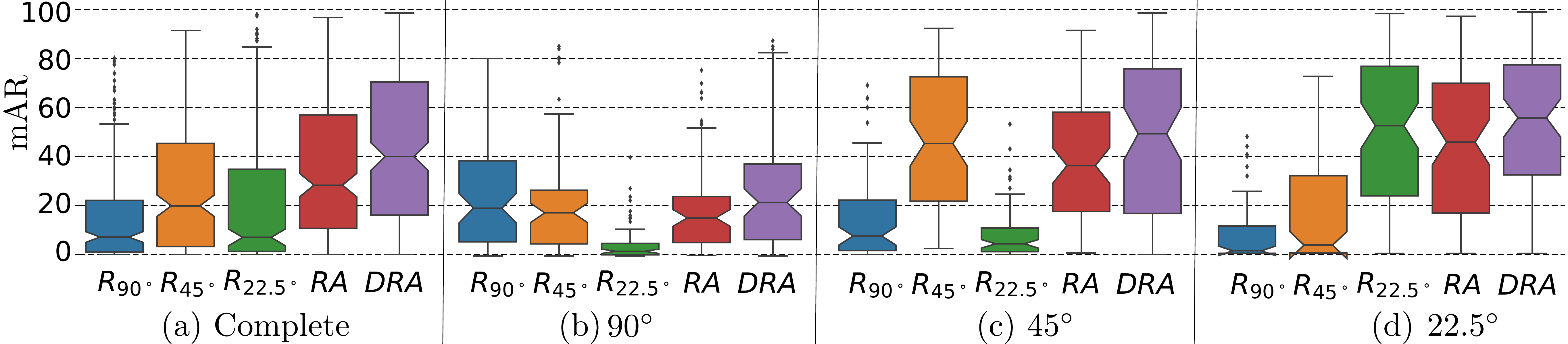}
    \caption{Distribution over Mean Average Recall.}
    \label{fig:mAR}
\end{figure*}

In ablation, we study the impact of the deflection metric and data augmentation, focusing on modifications in FoV. 
As our baselines, we use ResNet-50 with FPN, MaskRCNN, and PointRend as application head exclusively trained on a single vertical FoV $f_v \in \{22.5^\circ, 45^\circ, 90^\circ\}$ , which we further denote as $R_{22.5^\circ}$, $R_{45^\circ}$, and $R_{90^\circ}$ respectively. We also train this model with augmentation, further denoted as $RA$. Our model, as described in \autoref{sec_method} is further denoted as $DRA$ and also trained with augmentation. For the augmentation of $RA$ and $DRA$, we random change the $w \in \{512,1024,2048\}$, $h \in \{32,64,128\}$, and  $f_v \in \{22.5^\circ, 45^\circ, 90^\circ\}$ during training. For $R_{22.5^\circ}$, $R_{45^\circ}$, and $R_{90^\circ}$ we only augment $h$ and $w$. We refrain from training a model for each resolution to limit the scope of this work. For testing, we used the sequences from the test set and generated a small test set for each sequence at each resolution and field of view from $w$, $h$, and $f_v$, totaling 243 small test sets. We calculate the mAR for every subset. In \autoref{fig:mAR} the distribution of mAR over those small test sets for the considered models can be seen as box plots. We display the mAR over the complete dataset and exclusively for the subsets with $22.5^\circ$, $45^\circ$, and  $90^\circ$ FoV, respectively.
While there is some noticeable spread over mAR over the complete dataset for each model, we can see that the median mAR for the models $R_{22.5^\circ}$, $R_{45^\circ}$, and $R_{90^\circ}$ (marked as an indent in the box plots) is low, the median mAR for $RA$ is noticeably better, and the median mAR of $DRA$ is even better. This indicates a limitation of $R_{22.5^\circ}$, $R_{45^\circ}$, and $R_{90^\circ}$ to transfer to data captured from sensors with different FoVs. Furthermore, in the experiments in which we exclusively test the performance on a single FoV we can see that there are general differences in the mAR concerning the FoV.
We assume this is due to the size of the objects in pixel space. Objects appear smaller at a larger FoV, and the smaller an object is, the harder it is to detect it. 
The models that are exclusively trained on the evaluated FoV tend to work well, while those trained on other FoVs have a significantly lower mAR. It is noticeable that $RA$ seems to generalize but cannot catch up with the mAR of the models exclusively trained on the evaluated FoV. On the other hand, DRA achieves a slightly better mAR for each FoV under consideration compared to the models trained exclusively on the FoV under consideration.
\\

\textbf{Discussion}:

We showed in our experiment that using data from a single LiDAR sensor source for training might bias an instance segmentation model when applying the model to data from novel sensors with different FoVs. 
Using data from various sensors as simulated by the augmentation in $RA$ can increase the capability of a model to generalize. However, it is not possible to use the full potential of data from multiple sensors. The experiment shows that our method $DRA$ can help to generalize over various sensors without sacrificing quality compared to the sensor specific models.

We neglect some sensor effects by using simulated and idealized data. Nevertheless, we interpret our results so that our method can contribute to achieving sensor equivariance. Beyond our method, annotated data from multiple sensors and suitable augmentation strategies are needed for real world applications.

\section{\large Conclusions and Future Work}
\label{sec:conclution}

With this work, we presented a spherical projection model, a deflection metric, which can be used to encode geometric sensor properties to a projected image, and an architecture for joint processing of projected images and the deflection metric. The deflection metric is simple and thus efficient without incurring any additional computational requirements. In our experiment, we evaluated the usability of our method for the considered use case.

Since our method is suitable for processing LiDAR data and other sensors such as cameras, we would also like to consider other sensors and further use cases, as well as real-world data, to identify and relax the limitations of our approach. 

\section*{Acknowledgment}
This work results from the project KI Data Tooling supported by the Federal Ministry for Economic Affairs and Climate Action (BMWK), grant numbers 19A20001L and 19A20001O. Additionally, the work is supported by ``FuE Programm Informations- und Kommunikationstechnik Bayern'', grant number DIK-1910-0016// DIK0103/01.

{\small
\bibliographystyle{ieee_fullname}
\bibliography{main}
}

\end{document}